\documentclass{article}

\usepackage{PRIMEarxiv}

\usepackage[utf8]{inputenc} 
\usepackage[T1]{fontenc}    
\usepackage{hyperref}       
\usepackage{url}            
\usepackage{booktabs}       
\usepackage{amssymb}
\usepackage{amsthm}
\usepackage{amsmath}
\usepackage{amsfonts}       
\usepackage{nicefrac}       
\usepackage{microtype}      
\usepackage{lipsum}
\usepackage{fancyhdr}       
\usepackage{color}
\usepackage{subfigure}
\usepackage{graphicx}       
\graphicspath{{media/}}     

\newtheorem{assumption}{Assumption}

\newtheorem{theorem}{Theorem}
\newtheorem{remark}{Remark}
\newtheorem{lemma}{Lemma}

\newtheorem{corollary}{Corollary}

\title{Disturbance Estimation for High-Degree-of-Freedom Euler-Lagrangian Systems Using Sliding Mode Observer without Matching Conditions
\thanks{This work was supported by the Horizon 2020 program under grant 820742 of the project ``HR-Recycler''.} 
}

\author{
  Zengjie Zhang \\
  Department of Electronic Engineering \\
  Eindhoven University of Technology \\
  Eindhoven, Netherlands\\
  \texttt{z.zhang3@tue.nl} \\
   \And
  Dirk Wollherr \\
  Chair of Automatic Control Engineering \\
  Technical University of Munich \\
  Munich, Germany\\
  \texttt{dw@tum.de} \\
}

\begin{document}
\maketitle

\begin{abstract}
This paper proposes a novel observer-based disturbance estimation method for high degree-of-freedom Euler-Lagrangian systems using an unknown input-output (UIO) sliding mode observer (SMO). Different from the previous SMO methods, this approach does not assume the matching condition of the disturbances. Besides, compared to the conventional disturbance estimation methods, the proposed method does not require the calculation of the inverse inertia matrices and accurate measurement of system velocities. This advantage resolves the concerns of heavy computational load and amplified noise for practical problems like external torque estimation of high degree-of-freedom manipulators in safe human-robot collaboration. We achieve this by defining a novel linearized model for the Euler-Lagrangian system and designing a sliding-mode-based disturbance observer. The estimation precision of the observer is ensured by the Lyapunov-based stability proof and the equivalent control theory of sliding mode. At the end of this paper, the method is implemented on a seven-degree-of-freedom robot manipulator, and the experimental results confirm its decent performance and potential to be applied in practice.
\end{abstract}

\keywords{fault detection and isolation, disturbance estimation, Euler-Lagrangian system, safe human-robot collaboration, physical human-robot interaction, system observation, sliding mode observer.}

\definecolor{limegreen}{rgb}{0.2, 0.8, 0.2}
\definecolor{forestgreen}{rgb}{0.13, 0.55, 0.13}
\definecolor{greenhtml}{rgb}{0.0, 0.5, 0.0}

\section{Introduction}

Safety is always a critical issue in physical Human-Robot Interaction (HRI), of which an important aspect is to properly handle the accidental collisions between robots and the environment~\cite{flacco2013safe}. The concern of safe HRI motivates the investigations on the design of collision event handling pipelines~\cite{lasota2017survey, haddadin2017robot}. Related topics on collision handling include collision avoidance~\cite{
schafer2019safe}, collision force estimation~\cite{sun2017protective, popov2019real}, collision detection and identification~\cite{
kouris2018frequency} and collision reaction strategy design~\cite{
haddadin2008collision}. In this paper, we are concerned with the collision force estimation problem of robot systems, which is formulated as a disturbance or actuator fault estimation problem for Euler-Lagrangian systems. Among various existing approaches, we are especially interested in the observer-based disturbance estimation methods, or the analytical-redundancy methods~\cite{
de2008atlas, samy2011survey}, due to simplicity and high reliability. 

During the past two decades, various observer-based disturbance estimation schemes are proposed, including the transfer-function-based observers~\cite{
yun2013design, sariyildiz2014stability}, the intelligent-model-based methods~\cite{
li2014absolute, cho2019neural, jassim2019hybrid}, Luenberger observers~\cite{
gao2007actuator, jiang2006fault, zhu2015fault}, Nonlinear observers~\cite{
yang2016parameterization, nemati2019nonlinear}, sliding mode observers~\cite{yan2007nonlinear,
edwards2000sliding, brambilla2008fault, capisani2012manipulator, liu2013sensor, zhang2019integral}, the high gain observers~\cite{veluvolu2010nonlinear, boum2019high}. The transfer-function-based observers treat the Euler-Lagrangian system as a linear model by ignoring the coupled dynamics, producing less precise estimation results than other methods. The intelligent-model-based methods reconstruct the disturbances through training regressive models like neural networks or support vector machines, which apply only to state-dependent disturbances and usually require offline training processes. The Luenberger, high-gain and sliding mode observers, in comparison, are designed based on the feedback-linearization models of the Euler-Lagrangian systems. However, the conventional feedback-linearization scheme brings the inverse inertia matrix of the system to the disturbance gain. Therefore, the resulting disturbance observers require the inverse calculation of inertia matrices, which leads to tremendous computational loads and even ill conditions for Euler-Lagrangian systems with high degree-of-freedom (DoF). A typical example is the external torque estimation for high-DoF manipulators which are widely applied in safe-HRI tasks. The application of the conventional methods to such a problem is challenging, due to the strict real-time restrictions.

To overcome this challenge, a general-momentum observer is presented in~\cite{oh1999disturbance, de2003actuator}, where the Euler-Lagrangian system is transformed into a linear model by defining the general momentum as the state variable. Based on the linearized model, a linear observer is designed to reconstruct the disturbance. As a result, the estimation error dynamics of the observer is equivalent to a first-order filter, where the estimation error is the filtered output of the disturbance. Since this approach does not require the calculation of the inverse inertia matrix, it is widely applied to safe HRI systems and fault detection and isolation of robots~\cite{
de2006collision, hu2017contact, han2019collision}. However, this method requires the system velocity in the feedback loop. Actually, in most of the practical applications, the system velocity is not directly measurable but is obtained by differentiating the position measurement~\cite{floquet2007super}, which introduces noise. Including the velocity in the closed-loop amplifies the differential noise and affects the performance of the observer. Unfortunately, a disturbance estimation method that requires neither the inverse inertia matrix nor the velocity feedback still lacks.

Therefore, the main contribution of this paper is to fill this gap by presenting a disturbance estimation method without both the inverse inertia matrix and system velocity feedback. Specifically, we propose a novel linearized model for Euler-Lagrangian systems and design a sliding-mode-based observer for this model, which ensures accurate disturbance estimation. Note that it is a challenge that the linearized model formulates second-order dynamics, for which the conventional disturbance observation framework does not apply~\cite{edwards2000sliding, liu2013sensor}. We solve this problem with the partial dynamical collapse property of sliding mode and guarantee the estimation precision using the Lyapunov-based method and the equivalent control theory. The solution of feasible observer parameters is given by a linear matrix inequality (LMI). The remaining content of this paper is organized as follows. Sec. \ref{sec:main} interprets our main results, including the linearization scheme of the original system and the design of the disturbance observer. In Sec. \ref{sec:exper}, we validate our method on a seven-DoF robot manipulator platform by two experiments and evaluate its performance. Finally, Sec. \ref{sec:concl} concludes the paper.

\section{Main Results}\label{sec:main}

This section presents the main contribution of this paper. Firstly, we formulate the disturbance estimation problem for a Euler-Lagrangian system. Then, we transform the system into a linearized model and design a sliding mode disturbance estimator. Finally, we prove the error convergence of the observer and discuss its estimation accuracy. 

\subsection{Problem Formulation}

The dynamic model of an $n$-DoF rigid robot manipulator with external torques is represented as
\begin{equation}\label{eq:TorqueMod}
{M}({q})\ddot{{q}}+{N}({q},\dot{{q}})+{G}({q})+{F}\!\left( \dot{{q}}\right) = {\tau} + {\tau}_{\mathrm{d}},
\end{equation}
where $q(t), \dot{q}(t), \ddot{q}(t) \in \mathbb{R}^n$ are respectively the position, velocity and acceleration of the system in the joint space, ${M}({q})\in \mathbb{R}^{n\times n}$, ${N}({q},\dot{{q}})\in \mathbb{R}^{n}$, ${G}({q})\in \mathbb{R}^{n}$ and ${F} (\dot{{q}})\in \mathbb{R}^{n}$ are respectively the inertia matrix, the Coriolis and centrifugal vector,
the gravitational vector and the frictional vector, ${\tau} \in \mathbb{R}^{n}$ is the input torques applied to the robot actuators, and ${\tau}_{\mathrm{d}}(t) \in \mathbb{R}^{n}$ represents the external torques on the robot joints, which reflects disturbances, collision forces or actuator faults~\cite{de2003actuator}. Note that $\tau_{\mathrm{d}}(t)$ is time-dependent and free from the system states $q(t)$.
The system parameters are not perfectly known. Instead, only the identified parameters $\hat{M}(q)$, $\hat{N}(q, \bar{\dot{q}})$, $\hat{G}(q)$, $\hat{F}(\dot{q})$ are available. Additionally, we assume that the joint velocity $\dot{q}(t)$ is not measurable, but its approximation $\bar{\dot{q}}(t)$ is obtained by differentiating the position measurement $q(t)$, which leads to a bounded approximation error $\varepsilon_{\dot{q}}(t) \!=\! \dot{q}(t) - \bar{\dot{q}}(t)$, $\|\varepsilon_{\dot{q}}(t)\| \leq \epsilon_{\dot{q}}$, $\exists \, \epsilon_{\dot{q}} \in \mathbb{R}^+$. Based on these statements, we propose the following assumptions for this paper. Note that we respectively use $\|\cdot\|: \mathbb{R}^n\! \rightarrow\! \mathbb{R}^+$ and $\|\cdot\|_2: \mathbb{R}^{n \times n} \!\rightarrow\! \mathbb{R}^+$ to represent the 2-norms of vectors and matrices.

\begin{assumption}\label{as:as1}
The system velocity $\dot{q}(t)$, acceleration $\ddot{q}(t)$ and external torques $\tau_{\mathrm{d}}(t)$ of system (\ref{eq:TorqueMod}) are bounded, i.e., there exist $\alpha_1, \alpha_2, \alpha_{\tau} \in \mathbb{R}^+$, such that $\|\dot{q}(t)\| \leq \alpha_1$, $\|\ddot{q}(t)\| \leq \alpha_2$ and $\|\tau_{\mathrm{d}}(t)\| \leq \alpha_{\tau}(t)$ for all $t \in \mathbb{R}^+$.
\end{assumption}

\begin{assumption}\label{as:as2}
The identification errors of the system parameters are also bounded, i.e., $\|\tilde{M}(q)\|_2 \leq \epsilon_M$, $\|\tilde{N}(q, \dot{q}, \bar{\dot{q}})\| \leq \epsilon_N$, $\|\tilde{G}(q)\| \leq \epsilon_G$ and $\|\tilde{F}(q)\| \leq \epsilon_F$, $\epsilon_{M}, \epsilon_{N}, \epsilon_{G}, \epsilon_F \in \mathbb{R}^+$, where
\begin{equation*}
\begin{split}
\tilde{{M}}({q}) = {M}({q}) - \hat{{M}}({q}),&~\tilde{{N}}({q}, \dot{{q}}, \bar{\dot{{q}}}) = {N}({q},\dot{{q}}) - \hat{{N}}({q},\bar{\dot{{q}}}), \\
\tilde{{G}}({q}) = {G}({q})- \hat{{G}} ({q}) ,&~ \tilde{{F}}(\dot{{q}},\dot{\hat{{q}}}) = {F}(\dot{{q}}) - \hat{{F}}(\dot{\hat{{q}}}).
\end{split}
\end{equation*}
\end{assumption}

\begin{remark}
The boundedness of the system velocity and acceleration in Assumption \ref{as:as1} is based on the finitude of system energy and actuation~\cite{zhang2019integral}. Assumption \ref{as:as2} assumes that the system identification process is well performed. Both assumptions are frequently used in related literature. 
\end{remark}


The problem investigated in this paper is to reconstruct an estimation $\hat{\tau}_{\mathrm{d}}(t)$ for the disturbance $\tau_{\mathrm{d}}(t)$, which is formulated as an unknown-input observation problem. In practice, the online disturbance estimation methods based on analytical-redundancy provides an efficient and cheap solution for the safety of robot platforms with low-expense requirements. 

\subsection{System Transformation}

In this section, we transform the Euler-Lagrangian system (\ref{eq:TorqueMod}) to a linearized model  by defining two auxiliary variables $\zeta(q) = M(q)q$ and $\xi(q, \dot{q}) = M(q)\dot{q}$. In previous work, $\xi(q, \dot{q})$ is also referred to as the \textit{general momentum}~\cite{de2003actuator}. For brevity, we represent these variables as the time-dependent form $\zeta(t)$ and $\xi(t)$. From Assumption \ref{as:as1}, we know that $\xi(t)$ is bounded, i.e.,
\begin{equation*}
\|\xi(t)\| \leq \sup_{q \in \mathbb{R}^n}\|M(q)\|_2 \sup \|\dot{q}\| = \sigma_{\!M} \alpha_1,
\end{equation*}
where $\sigma_M \in \mathbb{R}^+$ is the spectral radius of $M(q)$, for which the boundedness of the inertia matrix $M(q)$ is applied (See~\cite{siciliano2016springer}, Page 134, Property 6.1). Notice that $\zeta(t)$ is measurable since it depends only on position measurement $q(t)$. Then, the time-derivatives of $\zeta(t)$ and $\xi(t)$ are
\begin{equation*}
\dot{\zeta} = \dot{\hat{M}}(q) q + \hat{M}(q) \dot{q}, ~\dot{\xi} = \dot{\hat{M}}(q) \dot{q} + \hat{M}(q) \ddot{q},
\end{equation*}
where $\dot{\hat{M}}(q) = \frac{\partial \hat{M}(q)}{\partial q} \dot{q}$ is the time derivative of the identified inertia matrix $\hat{M}(q)$, and $\dot{\xi}(t)$ further leads to
\begin{equation*}
\begin{split}
\dot{\xi} =& \, \dot{\hat{M}}(q) \dot{q} + M(q)\ddot{q} + \tilde{M}(q)\ddot{q} \\
=& \, \dot{\hat{M}}(q)\bar{\dot{q}} + \tau + \tau_{\mathrm{d}} - \hat{N}(q, \bar{\dot{q}}) \bar{\dot{q}} - \hat{G}(q) - \hat{F}(\bar{\dot{q}} ) + \eta (q,\dot{q}, \bar{\dot{q}} ),
\end{split}
\end{equation*}
where the uncertainty term $\eta$ reads
\begin{equation*}
\eta (q,\dot{q}, \bar{\dot{q}} ) \!=\! \left(\! \frac{\partial \hat{M}(q)}{\partial q} \dot{q} \! \right) \! \varepsilon_{\dot{q}} +\! \tilde{M}(q)\ddot{q}  - \tilde{N}(q,\dot{q}, \bar{\dot{q}}) - \tilde{G}(q) - \tilde{F}(\hat{\dot{q}}).
\end{equation*}
For brevity, we represent it as $\eta(t)$. From Assumption \ref{as:as1} and \ref{as:as2} we know that $\eta(t)$ is also bounded, i.e., $\|\eta(t)\| \leq \epsilon_{\eta}$, where
\begin{equation*}
\epsilon_{\eta} = \epsilon_{\dot{q}} \alpha_1 \! \sup_{q \in \mathbb{R}^n} \!\| {\partial \hat{M}(q)}/{\partial q} \|_2 + \epsilon_{M} \alpha_2  + \epsilon_N + \epsilon_G + \epsilon_F.
\end{equation*}

Thus, we formulate system (\ref{eq:TorqueMod}) as the following linear form,
\begin{equation}\label{eq:asystem}
	\begin{array}{rl}
		& \dot{{x}}(t) = {A} {x}(t) + {u}(t) + {E} {d}(t) \\
		& {\zeta}(t) = {C} {x}(t),
	\end{array}
\end{equation}
where $x(t) = \left[\,\zeta^{\!\top}\!(t) ~ \xi^{\!\top}\!(t) \, \right]^{\!\top}$ is the state variable, $A \in \mathbb{R}^{n \times n}$, $C \in \mathbb{R}^{n \times 2n}$, $E \in \mathbb{R}^{2n \times n}$ are parametric matrices formulated as
\begin{equation*}
	{A} = \left[\!\!
		\begin{array}{cc}
				0_n \!&\! {I}_n \\
				0_n \!&\!  0_n
		\end{array}
			\!\!\right],~
	{C} = \left[\!\!
		\begin{array}{cc}
			{I}_{n} \!&\! 0_{n}
		\end{array}
			\!\!\right],~
	{E} = \left[\!\!
		\begin{array}{c}
			0_n \\
			{I}_n
		\end{array}
			\!\!\right],
\end{equation*}
where $I_n,\,0_n \in \mathbb{R}^{n \times n}$ are respectively the $n$-dimensional identity and zero matrices, $\zeta(t)$ serves as the measurement output, $u(t)$ is the auxiliary input that reads
\begin{equation*}
	{u} = \left[\!\! \begin{array}{c}
	\dot{\hat{{M}}}({q}) {q} \\
	{\tau}  + \dot{\hat{{M}}}({q}) \bar{\dot{{q}}} - \hat{{N}}({q},\bar{\dot{{q}}}) - \hat{{G}}({q})- \hat{{F}}(\bar{\dot{{q}}})
	\end{array} \!\!\right],
\end{equation*}
and the disturbance term $d(t) = \tau_{\mathrm{d}}(t) + \eta(t)$ contains both the unknown external torques $\tau_{\mathrm{d}}(t)$ and the unmodeled system uncertainties $\eta(t)$. Note that $d(t)$ is also bounded, i.e., $\|d(t)\| \leq \alpha_{\tau} + \epsilon_{\eta}$, and we claim $d \approx \tau_{\mathrm{d}}$, if $\epsilon_{\eta} \ll \alpha_{\tau}$. Under this condition, reconstruction of the unknown input $d(t)$ in the linearized model (\ref{eq:asystem}) is equivalent to the estimation of the disturbance $\tau_{\mathrm{d}}(t)$ of system (\ref{eq:TorqueMod}). To ensure the feasibility of the observer-based solution for the disturbance estimation problem, we present the following lemma.

\begin{lemma}\label{lm:lem}
	For the linearized dynamics in (\ref{eq:asystem}), $(A,E)$ is controllable and $(A,C)$ is observable.
\end{lemma}

\begin{proof}
	Calculating the controllability matrix $\left[\,
		{E}~
		{A} {E}
	\,\right]$ and the observability matrix $\left[\,
		{C}^{\top}~
		({C} {A})^{\top}
	\,\right]^{\top\!}$ for (\ref{eq:asystem}), we obtain
	\begin{equation*}
		\mathrm{rank} \left[\!\! \begin{array}{c}
		{C} \\
		{C} {A}
	\end{array} \!\!\right] = \mathrm{rank} \left[\!\! \begin{array}{cc}
		{I}_n \!&\! {0}_n \\
		{0}_n \!&\! {I}_n
	\end{array} \!\!\right] = 2n,
	\end{equation*}
	and
	\begin{equation*}
		\mathrm{rank} \left[\!\! \begin{array}{cc}
		E \!&\! AE
	\end{array} \!\!\right] = \mathrm{rank} \left[\!\! \begin{array}{cc}
		{0}_n \!&\! {I}_n \\
		{I}_n \!&\! {0}_n
	\end{array} \!\!\right] = 2n.
	\end{equation*}
	Thus, $(A,E)$ is controllable and $( {A}, {C} )$ is observable.
\end{proof}
Lemma \ref{lm:lem} indicates that it is feasible to design a disturbance observer that uses the measurement output $\zeta(t)$ to reconstruct the unknown input $d(t)$, which justifies the effectiveness of the linearized model (\ref{eq:asystem}). Different from the linearized model applied in the conventional observer-based methods, the disturbance gain matrix $E$ of (\ref{eq:asystem}) is constant and does not depend on the inverse inertia matrix $M^{\!-1\!}(q)$. However, it is noticed that, $CE = 0$ always holds, which means that the \textit{canonical form} of the linearized model (\ref{eq:asystem}) required in the conventional disturbance estimation framework~\cite{edwards2000sliding} does not exist. This is because the measurement output $\zeta(t)$ and the unknown input $d(t)$ formulate a second-order system. Therefore, reconstructing $d(t)$ from $\zeta(t)$ requires a second-order observer, such as the super-twisting sliding mode observer~\cite{floquet2007super, capisani2012manipulator}. Nevertheless, this method suffers from the lack of rigorous stability proofs and difficult application to multi-input multi-output systems. 
In the next section, we propose a novel observer design scheme for  (\ref{eq:asystem}), of which the estimation accuracy is guaranteed by the Lyapunov-based method and the equivalent control theory.

\subsection{Observer Design}\label{sec:obde}
For the linearized model (\ref{eq:asystem}), we design a state-disturbance observer as follows,
\begin{equation}\label{eq:observer}
\begin{split}
	\dot{\hat{{x}}}(t) =& {A} \hat{{x}}(t) + {u}(t) + {L}\! \left({\zeta}(t) - {C} \hat{{x}} (t)\right) + {K} {v}\!\left({s}\right),\\
	\hat{d}(t) =& -v_{\mathrm{eq}}(t), 
\end{split}
\end{equation}
where $\hat{d}(t)$ is disturbance estimation, $\hat{x}(t) = [\,\hat{\zeta}^{\top}\!(t)~\hat{\xi}^{\top}\!(t)\, ]^{\top}$ is state estimation which brings up the estimation error $e(t) = x(t) - \hat{x}(t)$, $e_{\zeta}(t) = \zeta(t) - \hat{\zeta}(t)$ and $e_{\xi}(t) = \xi(t) - \hat{\xi}(t)$, $s(t) = HC e(t)$ is the sliding mode variable, $H \in \mathbb{R}^{n \times n}$, $L, K \in \mathbb{R}^{2n \times n}$ are constant matrices to be determined, $v(s)$ is a switching control input that reads
\begin{equation}\label{eq:vseq}
v(s) = - \varrho(e_{\zeta},\hat{\xi}) \frac{s(t)}{\|s(t)\|},~\|s(t)\| \neq 0,
\end{equation}
where $\varrho(e_{\zeta}, \hat{\xi}) > 0$ is a positive definite scalar function of $e_{\zeta}(t)$ and $\hat{\xi}(t)$ to be designed later, and $v_{\mathrm{eq}}(t) \in \mathbb{R}^n$ is the equivalent control of $v(s)$ in the Filippov sense~\cite{haskara1998sliding}, which is to be discussed in Sec. \ref{sec:dest}. It is noticed that the observer (\ref{eq:observer}) does not contain the inverse inertia matrix $M^{\!-1\!}(q)$. Additionally, the approximated velocity $\bar{\dot{q}}(t)$ does not show up as a feedback, but only appear in the feedforward compensation term $u(t)$. As a result, $\varepsilon_{\dot{q}}(t)$ is not amplified in the closed-loop, but instead, attenuated by the feedback gain $L$ in the feedforward loop. These two properties of the observer (\ref{eq:observer}) reveal its advantages compared to the conventional methods. 

The selection of matrix $L$ is the main focus in this section. By calculating (\ref{eq:asystem}) $-$ (\ref{eq:observer}), we obtain the following estimation error dynamics,
\begin{equation}\label{eq:errordy}
	\dot{{e}}(t) = A_L {e}(t) + {E} d(t) - {K}{v} ({s}).
\end{equation}
where $A_L = {A} - {L} {C}$. Assuming that $Ed(t) - Kv(s) \equiv 0$, the global asymptotic stability of (\ref{eq:errordy}) at zero equilibrium $e(t) =0$ requires $A_L$ to be Hurwitz. Moreover, the eigenvalues of $A_L$ should also be confined to achieve fast convergence for the estimation errors. Thus, we require
\begin{equation}\label{eq:eigenk}
	\mathrm{Re}\!\left( \lambda_i(A_L) \right) < -\kappa,~i=1,2,\cdots,2n,
\end{equation}
where $\kappa \in \mathbb{R}^+$ is a predefined positive constant scalar, $\lambda_i(A_L)$ is the $i$-th eigenvalue of $A_L$, and $\mathrm{Re}(\cdot)$ denotes the real part of a complex. The condition (\ref{eq:eigenk}) means that all the eigenvalues of $A_L$ are located on the left of $-\kappa$, which ensures $e(t)$ a convergence rate faster than $e^{-\kappa t}$. Nevertheless, when the unknown input $d(t)$ is not zero, its influence on $e(t)$ is indicated by the following $H_\infty$ criteria,
\begin{equation}\label{eq:hinf}
	{H}_{d}^e = \sup_{\|{d}\| \neq 0} \frac{\left\| {e}(t) \right\|_2}{\left\| {d}(t) \right\|_2}.
\end{equation}
To reduce the influence of $d(t)$, we require ${H}_d^e < \gamma$ for a predefined constant scalar $\gamma >0$. Therefore, the solution of $L$ resolving these concerns is provided by the following Theorem.

\begin{theorem}\label{th:theo1}
If for given $\kappa, \gamma \in \mathbb{R}^+$, there exist ${P} \in \mathbb{R}^{2n \times 2n}$ and ${W} \in \mathbb{R}^{2n \times n}$, where $P$ is symmetrically positive definite, such that the following LMI holds,
\begin{equation}\label{eq:inequa1}
		\left[\!\!
	\begin{array}{cc}
		{\Lambda} \!&\! {PE}\\
		{E}^{\top}\!P \!&\! -\gamma^2 {I}
	\!\!\!\end{array}\right] < 0 ,
\end{equation}
where $\Lambda = {P}(\kappa {I} +{A})+ (\kappa {I} +{A})^{\top}\!{P} +{I} -{WC}-{C}^{\top}\!{W}^{\top}\!$, and ${L} = {P}^{-1}{W}$, then for the error dynamics (\ref{eq:errordy}), the following conditions are satisfied.

(a). Condition (\ref{eq:eigenk}) holds,

(b). For any $\|d(t)\| \neq 0$, the $H_{\infty}$ criteria ${H}_{d}^e < \gamma$ holds.

\end{theorem}

\begin{proof}
	(a). Substituting $L=P^{-1}W$ to (\ref{eq:inequa1}), and using property of Schur complement in Lemma \ref{lm:lem2}, the LMI in (\ref{eq:inequa1}) leads to
		\begin{equation}\label{eq:riccati}
			\begin{split}
				{P}(\kappa {I} +{A} - {L}{C}) &+ (\kappa {I} +{A} -{L}{C})^{\top}\!{P} \\
				&+ {I}+ \frac{1}{\gamma^2} {PE}({PE})^{\top}\! < 0.
			\end{split}
		\end{equation}
	Considering ${PE}({PE})^{\top}\! \geq 0$, from (\ref{eq:riccati}), we have
	\begin{equation*}
		{P}(\kappa {I} +{A} -{LC})+ (\kappa {I} +{A} -{LC})^{\top}\!{P} < -{I},
	\end{equation*}
	which indicates that $\kappa {I} +{A} -{LC}$ is Hurwitz and its eigenvalues satisfy $\mathrm{Re}\left(\lambda_i\!\left(\kappa {I} +{A} -{LC}\right)\right) < 0$, for all $i = 1,2,\cdots,2n$. This straight-forwardly leads to (\ref{eq:eigenk}).

	(b). Let us consider the linear dynamics (\ref{eq:asystem}) in the frequency domain. From (\ref{eq:riccati}), we know
	\begin{equation}\label{eq:kp}
		\begin{split}
			 -{P}(j \omega {I} -{A} +{L}{C})&- (j \omega {I} -{A} +{L}{C})^{\top}\!{P} + {I}\\
			&+ \frac{1}{\gamma^2} {P}{E}({PE})^{\top}\! < -2 \kappa {P}.
		\end{split}
	\end{equation}
	where $j\omega$ represents the Laplace operator. Then, $e(t)$ and $d(t)$ are represented as $e(j\omega)$ and $d(j\omega)$ in the frequency domain, and the transfer function of the linear dynamics (\ref{eq:asystem}) reads
	\begin{equation*}
	e(j \omega) = \left( j \omega {I} - {A} +{LC} \right)^{-1} \!E {d}(j \omega).
	\end{equation*}
	Multiplying $e^{\!\top}\!(j\omega)$ and $e(j \omega)$ on the left and right sides of (\ref{eq:kp}), we obtain 
\begin{equation}\label{eq:kp_2}
	\begin{split}
		&\|\left( j \omega {I} - {A} + {LC} \right)^{-1} \!\!Ed(j \omega)\|^2 - \gamma^2 \| {d}(j \omega) \|^2\\
		&+\left\| \left( \frac{1}{\gamma} {E}^{\top}\!{P}^{\top}\! \left( j \omega {I} - {A} + {LC} \right)^{-1}\!\!E + \gamma {I} \right)\! d(j \omega) \right\|^2 \\
		& < - 2 \kappa \left\| {L}_p \left( j \omega {I} - {A} + {LC} \right)^{-1}\!\! Ed (j \omega) \right\|^2,
	\end{split}
\end{equation}
where ${L}_p^{\top}\!{L}_p = {P}$ is the Cholesky decomposition of ${P}$. Thus, from (\ref{eq:kp_2}), we have 
\begin{equation}\label{eq:kp_3}
	\|\left( j \omega {I} - {A} + {LC} \right)^{-1}\!\! {E} {d}(j \omega)\|^2 < \gamma^2 \| {d} (j \omega)\|^2,
\end{equation}
which leads to
\begin{equation*}
\sup_{\|d\| \neq 0} \frac{\left\| {e}(j \omega) \right\|}{\left\| {d}(j \omega) \right\|} < \gamma,
\end{equation*}
and $H_d^e < \gamma$ holds, due to the equivalence between the time and frequency domains.
\end{proof}

\begin{remark}
Theorem \ref{th:theo1} ensures that $e(t)$ is uniformly ultimately bounded for $d(t) \neq 0$, i.e., $\|e(t)\| \leq \gamma \|d(t)\|$. The restriction on the eigenvalues in (\ref{eq:eigenk}) confines a minimal convergence rate $e^{-\kappa t}$. The boundedness of $e(t)$, ensured by Theorem \ref{th:theo1}, is important to ensure a stable observation. Note that (\ref{eq:inequa1}) is not necessarily feasible for arbitrary $\kappa$ and $\gamma$. The balance between these two parameters reflects the compromise between the convergence rate and the ultimate boundary of $e(t)$.
\end{remark}

Before we introduce the determination of the other parameters and the estimation law $\hat{d}(t)$ of observer (\ref{eq:observer}), we give the following properties of the matrix $P$ in Theorem \ref{th:theo1}, which is frequently used in the following section and the appended proofs.

\begin{lemma}\label{lm:lem2}
The Schur compliment property (\cite{zhang2006schur}, Theorem 1.12): 
let $P \in \mathbb{R}^{2n \times 2n}$ be a symmetrical real matrix that is partitioned as 
\begin{equation}\label{eq:p_decom}
	{P} = \left[\!\!\begin{array}{cc}
		{P}_{11} \!&\! {P}_{12}\\
		{P}_{12}^{\top}\! \!&\! {P}_{22}
	\end{array}\!\!\right],
\end{equation}
where $P_{11}, P_{12}, P_{22} \in \mathbb{R}^{n \times n}$ and $P_{11}$ is non-singular. Then $P>0$ if and only if $P_{11} > 0$ and $P_{22} - P_{12}^{\top}P_{11}^{-1}\! P_{12} >0$.
\end{lemma}

\begin{corollary}\label{cr:coro}
If there exists a symmetrically positive definite $P \in \mathbb{R}^{2n \times 2n}$, such that (\ref{eq:inequa1}) holds, and $P$ is partitioned as (\ref{eq:p_decom}), then ${P}_{11} > 0$, ${P}_{22} > 0$, and ${P}_{12}$ is non-singular and Hurwitz.
\end{corollary}

\begin{proof}
From (\ref{eq:p_decom}), since $P$ is symmetrically positive definite and $P_{11}$ is square, we know ${P}_{11}>0$. Therefore, according to Lemma \ref{lm:lem2}, we have 
\begin{equation}\label{eq:pschur}
P_{22} - P_{12}^{\top}P_{11}^{-1} P_{12}^{\top} >0.
\end{equation}
Considering $P_{12}^{\!\top}P_{11}^{-1} P_{12}^{\!\top} \geq 0$, (\ref{eq:pschur}) leads to $P_{22} > 0$. In the meantime, the LMI (\ref{eq:riccati}) leads to 
	\begin{equation}\label{eq:new_ric}
		{P}({A - {L}{C}}) + ({A} - {L}{C})^{\top}\!{P} + {I} < 0.
	\end{equation}	
	We substitute the partitioned matrix $P$ in (\ref{eq:p_decom}) to (\ref{eq:new_ric}), and represent the result in the following partitioned form,
	\begin{equation}\label{eq:pp_decom}
		\tilde{P} = \left[\!\!\begin{array}{cc}
		{\tilde{P}}_{11} & {\tilde{P}}_{12} \\
		{\tilde{P}}_{12}^{\top}\! & {\tilde{P}}_{22}
	\end{array}\!\!\right] > 0,
	\end{equation}
	where $\tilde{P} = - {P}({A - {L}{C}}) - ({A} - {L}{C})^{\top}\!{P} - {I}$, ${\tilde{P}}_{11} = P_UL + (P_UL)^{\!\top} - {I}$, ${\tilde{P}}_{12} = {L}^{\!\top}\!{P}_{R} - {P}_{11}$ and ${\tilde{P}}_{22} = - {P}_{12} - {P}_{12}^{\top}\! - {I}$, where $P_U = [\, {P}_{11} ~ {P}_{12} \,]$ and $P_R = [\, {P}_{12}^{\top} ~ {P}_{22}^{\top} \,]^{\top}$ are respectively the upper and right partitions of $P$. Since $\tilde{P} >0$ and $\tilde{P}_{11}$ is square, similar to $P$, we have
\begin{equation}\label{eq:eq1}
	{\tilde{P}}_{22} = - {P}_{12} - {P}_{12}^{\top}\! - {I} > 0,
\end{equation}
which indicates that $P_{12}$ is non-singular and Hurwitz.
\end{proof}	

\subsection{Disturbance Estimation}\label{sec:dest}

The precision of the disturbance estimation $\hat{d}(t)$ in (\ref{eq:observer}) is based on the finite-time convergence of the sliding variable $s(t)$, for which we have the following theorem.

\begin{theorem}\label{th:theo2}
	If for given $\kappa, \gamma \in \mathbb{R}^+$, there exist $P\in \mathbb{R}^{2n \times 2n}$ and $W \in \mathbb{R}^{2n \times n}$, where $P$ is symmetrically positive definite, such that all conditions in Theorem 1 are satisfied, $H,\,K$ in (\ref{eq:observer}) are selected as $H \!=\! -P_{22} P_{12}^{-1} P_{11} \!+\! P_{12}^{\top}$, $K \!=\![\,K_0^{\top}~I_n\,]^{\top}$, where $K_0 \!=\! -\!\left( P_{12}^{-1}\right)^{\!\!\top} \!\!P_{22}$, then the following conditions hold.
	
(a). The matrices $H$ is non-singular and $P_K\!=\! K^{\!\top}\!PK$ is symmetrically positive definite.
	
(b). If $\varrho(e_{\zeta}, \hat{\xi})$ in (\ref{eq:vseq}) is determined as 
\begin{equation}\label{eq:rhogain}
\begin{split}
\varrho(e_{\zeta}, \hat{\xi}) =& \, \varrho_0 + \|Q_{\zeta} e_{\zeta}\| + \|Q_{\xi} \hat{\xi}\| + \sigma_{\!M} \alpha_1 \|Q_{\xi}\|_2 \\
& + (\alpha_{\tau} + \epsilon_{\eta} ) \|Q_{d}\|_2, 
\end{split}
\end{equation}
where $\varrho_0 \in \mathbb{R}^+$ is a predefined constant positive scalar, $Q_{\zeta} = P_K^{-1}K^{\!\top}\!P A_L^L$, $Q_{\xi} = P_K^{-1}K^{\!\top}\!P A_L^R$, $Q_{d} = P_K^{-1}K^{\!\top}\!PE$, where $A_L^L,A_L^R \in \mathbb{R}^{2n \times n}$ are partitions of $A_L$, i.e., $A_L = \left[\, A_L^L~A_L^R\, \right]$, then for any initial condition $s(0) = s_0 \in \mathbb{R}^n$, $s(t)$ reaches zero in a finite time $t_0 \in \mathbb{R}^+$, $t_0 < {\|s_0\|}/{(\varrho_0 \lambda_{\min}(P_K)}$, where $\lambda_{\min}(P_K))$ is the minimal eigenvalue of $P_K$.
\end{theorem}

\begin{proof}
(a). It is straight-forward to verify that
\begin{equation*}
K^{\!\top}\!P = HC = \left[\, -P_{22} P_{12}^{-1} P_{11} \!+\! P_{12}^{\top} ~~0 \,\right],
\end{equation*}
which leads to
\begin{equation}\label{eq:pkk}
\begin{split}
K^{\!\top}\!PK  =& \, P_{22} P_{12}^{-1}\! P_{11}\! \left(P_{12}^{-1} \right)^{\!\top}\!\! P_{22}  -P_{22} \\
=& \, P_{22} P_{12}^{-1}\! \left( P_{11} - P_{12}P_{22}^{-1}P_{12}^{\!\top} \right)\! \left(P_{12}^{-1} \right)^{\!\!\top}\!\! P_{22}.
\end{split}
\end{equation}
Note that the existence of $P_{12}^{-1}$ is ensured by Corollary \ref{cr:coro}. According to the property of the Schur compliment in Lemma \ref{lm:lem2}, we know $P_{11} - P_{12}P_{22}^{-1}P_{12}^{\!\top} > 0$, since $P>0$ and $P_{22} > 0$. Therefore, (\ref{eq:pkk}) indicates that $P_K = K^{\!\top}\!PK$ is symmetrically positive definite.

(b). The derivative of the switching variable $s(t)$ reads
\begin{equation*}
	\dot{{s}}(t) = {HC}\dot{{e}}(t) = {K}^{\!\top}\!{P}\dot{{e}}(t).
\end{equation*}
Since $P_K$ is symmetrically positive definite (Theorem \ref{th:theo2}, (a)), we define the following Lyapunov function
\begin{equation*}
	V(t) = \frac{1}{2} {s}^{\top}\!(t) P_K^{-1} {s}(t),
\end{equation*}
which is positive definite for all $s(t) \neq 0$, 
and its derivative reads
\begin{equation}\label{eq:v_sd}
\begin{split}
	\dot{V} =& \, {s}^{\!\top}\! P_K^{-1} \dot{{s}} 
	= {s}^{\!\top}\! P_K^{-1} {K}^{\top}\!{P} \dot{{e}} \\
	=& \, {s}^{\!\top}\! P_K^{-1} {K}^{\top}\! {P} \!\left( A_L {e} + {K} {v}(s) -{E} {d} \right) \\
	=& \, {s}^{\!\top}\! P_K^{-1} {K}^{\top}\! {P} \!\left( A_L^L {e}_{\zeta} + A_L^R e_{\xi} + {K} {v}(s) -{E} {d} \right) \\
	\leq & \, \|{s}\| (\|Q_{\zeta} {e}_{\zeta} \| \!+\!\|Q_{\xi} {e}_{\xi} \|) \!-\! \varrho(e_{\zeta},\hat{\xi}) \|{s}\| \!+\!\|{s}\| \|Q_d {d} \|. 
\end{split}	
\end{equation}	
From the boundedness of $d(t)$ and $\xi(t)$, we have $\|Q_E {d} \| \leq (\alpha_{\tau} + \epsilon_{\eta}  ) \|Q_d\|_2$ and $\|Q_{\xi}e_{\xi}\| = \| Q_{\xi}\xi- Q_{\xi}\hat{\xi}\| \leq \|Q_{\xi}\|_2\|\xi\| + \|Q_{\xi}\hat{\xi} \| \leq \sigma_{\!M} \alpha_1 \|Q_{\xi}\|_2 + \|Q_{\xi}\hat{\xi} \|$. Substituting these inequalities to (\ref{eq:v_sd}), we obtain
\begin{equation*}
\begin{split}
\dot{V}  \leq & \, \|s\|\|Q_{\zeta} e_{\zeta}\| + \|s\|\|Q_{\xi} \hat{\xi}\| + \sigma_{\!M} \alpha_1 \|Q_{\xi}\|_2 \|s\| \\
& + (\alpha_{\tau} + \epsilon_{\eta} ) \|Q_d\|_2 \|s\| - \varrho(e_{\zeta}, \hat{\xi})\|s\|.
\end{split}
\end{equation*}
Substituting (\ref{eq:rhogain}) to it, we obtain
\begin{equation*}
\dot{V} = - \varrho_0 \|s\| \leq - \varrho_0 \sqrt{\frac{2V}{\lambda_{\mathrm{max}}\!\left( P_K^{-1} \right)}},
\end{equation*}
where $\lambda_{\max}(P_K^{-1})$ is the maximal eigenvalue of $P_K^{-1}$. We define a positive semi-definite function $V_s(t) \geq 0$, where $\dot{V}_s(t) = - \varrho_0 \sqrt{2 V_s(t)/\lambda_{\max}( P_K^{-1})}$, and $V_s(0) = V(0) = s_0^{\!\top}P_K^{-1}s_0/2$, which corresponds to the following solution,
\begin{equation}\label{eq:fivs}
V_s(t) \!=\! \left\{\!\!\! \begin{array}{ll}
\displaystyle \left(\! \sqrt{V(0)} \!-\! {\varrho_0 t}/{\sqrt{2 \lambda_{\max}( P_K^{-1})}}   \right)^{\!\!2}\!, \!\!\!&\!\! 0 \leq t \leq t_s \\
0,\!\!\! &\!\! t \geq t_s,
\end{array} \right.
\end{equation}
where 
\begin{equation*}
t_s = \frac{1}{\varrho_0} \sqrt{s_0^{\!\top}P_K^{-1}s_0 \lambda_{\max}\!\left( P_K^{-1} \right)} \leq \frac{\|s_0\|}{\varrho_0 \lambda_{\min}(P_K)} .
\end{equation*}
Note that the solutions $V(t)$ and $V_s(t)$ are unique in the Filippov sense (See~\cite{utkin1999sliding}, Sec. 3.5), and the comparison principle (See~\cite{khalil2002nonlinear}, Sec. 3.4) indicates that $0 \leq V(t) \leq V_s(t)$, $\forall t \in \mathbb{R}^+$. Therefore, $V_s(t)$ converges to zero within a finite time $t_s$, according to (\ref{eq:fivs}), and $V(t)$ converges to zero within a shorter time $t_0 \leq t_s \leq \|s_0\|(\varrho_0 \lambda_{\min}(P_K))$.
\end{proof}

\begin{remark}
Theorem \ref{th:theo2} indicates that the sliding mode variable $s(t)$ converges to zero within a finite time $t_0$ for arbitrary initial conditions. Specifically, a larger $\varrho_0$ leads to a shorter $t_0$. Meanwhile, considering the definition of $s(t)$ in Sec. \ref{sec:obde}, we know $s(t) = He_{\zeta}(t)$. Since $H$ is non-singular (Theorem \ref{th:theo2}, (a)), $e_{\zeta}(t)$ also reaches zero within time $t_0$.
\end{remark}

After $s(t)$ or $e_{\zeta}(t)$ reach zero, the following \textit{dynamical collapse} condition holds in the Filippov sense~\cite{shtessel2014introduction},
\begin{equation}\label{eq:slmc}
e_{\zeta}(t) =  \dot{e}_{\zeta}(t) = 0, t> t_0.
\end{equation}
Let us partition ${L}$ as $L = \left[\, {L}_1^{\top}~{L}_2^{\top} \, \right]^{\!\top}$, where ${L}_1, {L}_2 \in \mathbb{R}^{n \times n}$, and substitute it to $A_L$. Then, we obtain the partitioned form of $A_L$,
\begin{equation*}
A_L = A - LC = \left[\!\! \begin{array}{cc}
-L_1 \!&\! I \\
-L_2 \!&\! 0
\end{array} \!\!\right].
\end{equation*}
Thus, we rewrite the error dynamics (\ref{eq:errordy}) as the following partitioned form, 
\begin{equation}\label{eq:errpar}
\begin{split}
\dot{e}_{\zeta}(t) =& \, -L_1 e_{\zeta}(t) + e_{\xi}(t) + K_0 v(s) \\
\dot{e}_{\xi}(t) =& \, -L_2 e_{\zeta}(t) + v(s) + d(t).
\end{split}
\end{equation}
Note that $v(s) = v_{\mathrm{eq}}(t)$ holds in the Filippov sense. Then, substituting $v_{\mathrm{eq}}(t)$ and the sliding mode condition (\ref{eq:slmc}) to (\ref{eq:errpar}), we obtain the following dynamics,
\begin{equation}\label{eq:fil}
\dot{v}_{\mathrm{eq}}(t) = - K_0^{-1} v_{\mathrm{eq}}(t) - K_0^{-1} d(t).
\end{equation}
Since $P_{12}$ is Hurwitz and $P_{22}>0$ (Corollary \ref{cr:coro}), $-K_0^{-1} = P_{22}^{-1} P_{12}^{\top}$ is also Hurwitz. Therefore, the dynamics (\ref{eq:fil}) represents a first-order low-pass filter, where $v_{\mathrm{eq}}(t)$ is the filtered output of $d(t)$. If the bandwidth of the filter, which is determined by $K_0^{-1}$, is sufficiently large to cover $d(t)$, then we have $v_{\mathrm{eq}}(t) \approx -d(t)$ in the steady state. Therefore, $\hat{d}(t) = - v_{\mathrm{eq}}(t)$ in (\ref{eq:observer}) formulates an accurate estimation of $d(t)$. In practical applications, $v_{eq}(t)$ can be obtained by attaching a low-pass filter to the switching term $v(s)$~\cite{utkin1999sliding}.

Most of the conventional observer-based methods achieve an accurate disturbance estimation utilizing their equivalence to first-order filters, which also explains why they do not apply to second-order systems like (\ref{eq:asystem}). In this paper, we reduce the linearized model (\ref{eq:asystem}) to a first-order system (\ref{eq:fil}) by forcing \textit{partial dynamical collapse} using the switching input $v(s)$. To attenuate the chattering phenomenon, we apply the boundary-layer method to $v(s)$ by modifying it as
\begin{equation*}
v(s) = -\varrho(e_{\zeta}, \hat{\xi}) \frac{s(t)}{\|s(t)\| + \delta_s},
\end{equation*}
where $\delta_s \in \mathbb{R}^+$ is a properly selected boundary-layer scalar. As a result, the dynamical collapse condition in (\ref{eq:slmc}) does not strictly hold, which leads to steady-state errors to the disturbance estimation. Nevertheless, as long as the boundary-layer scalar is sufficiently small as $\delta_s \ll \alpha_{\tau}$, the steady-state errors can be neglected.

\section{Experiment}\label{sec:exper}

\begin{figure}[htbp]
\centering
\includegraphics[height=0.48\textwidth]{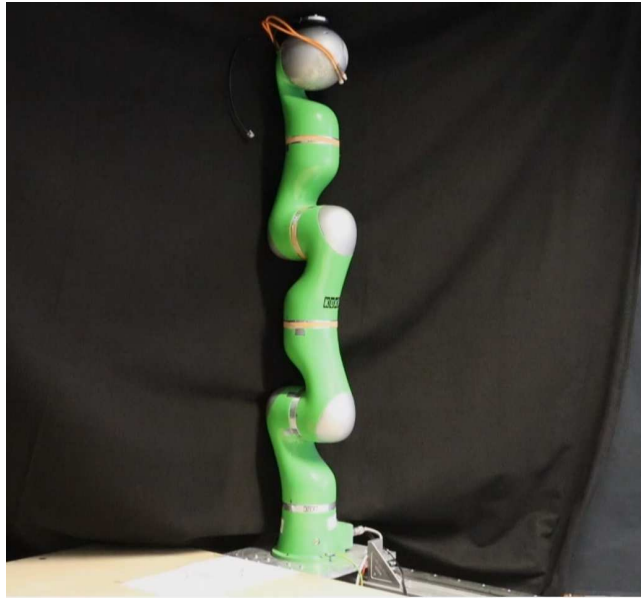}
\caption{The 7-DoF Kuka LWR 4+ Robot Manipulator.}
\label{fig:robot}
\end{figure}

In this section, we validate the proposed disturbance estimation method on a seven-DoF Kuka LWR 4+ robot manipulator which is designed for human-collaborative tasks~\cite{bischoff2010kuka}, as shown in Fig.~\ref{fig:robot}. The manipulator device provides a Fast Research Interface (FRI)~\cite{schreiber2010fast} for the design of robot controllers, which enables torque-level control capabilities, joint position feedback, system parameters acquisition and external torque measurements. To evaluate the performance of the proposed method for external torque estimation in nominal robot tasks, we conduct two experiments. (a). Insert three different types of predefined disturbance torques to the actuation input, and compare them with the estimated results; (b). Exert physical contacts on the robot end-effector and compare the estimated results with the signal data measured by the joint torque sensors. To simulate a working environment for the robot, we design a periodic reference trajectory $q_{\mathrm{r}}(t)$ in the joint space, for both experiments,
\begin{equation*}
q_{\mathrm{r}}(t) = \left( 1-\cos\!\left({\pi t}/{4}-1 \right) \right) \!q_{\mathrm{t}},~4 < t \leq 92,
\end{equation*}
where $q_{\mathrm{t}} = [~0.7~~0.7~~0.7~~0.7~~0.7~~0.7~~0.7\,]^{\top}\,$rad, and $q_{\mathrm{r}}(t) = 0$ for $t \leq 4$ and $t > 92$. It can be verified that $q_{\mathrm{r}}(t)$ and $\dot{q}_{\mathrm{r}}(t)$ are continuously bounded, and $\ddot{q}_{\mathrm{r}}(t)$ is bounded. A PD tracking controller $\tau(q,\dot{q})$ is designed for $\dot{q}_{\mathrm{r}}(t)$,
\begin{equation*}
\tau = K_{\mathrm{p}}\!\left(q - q_r \right) + K_{\mathrm{\mathrm{d}}}\!\left( \dot{q} - \dot{q}_r \right),
\end{equation*}
where $K_{\mathrm{p}} = 200 I_7$ and $K_{\mathrm{d}} = 8 I_7$. All programs are created by Simulink of MATLAB 2017a in Ubuntu 14.04 LTS and compiled using FRI. The sampling rate of the system is $1\,$kHz. Each experiment starts from the zero initial condition $q(0) = \dot{q}(0) = 0$ and lasts for 100 seconds. As a result, the linearized dynamics (\ref{eq:asystem}) is also given a zero initial condition $x(0)=0$.

\subsection{Experiment 1: Predefined Disturbances}
\begin{figure}[htbp]
\centering
\includegraphics[width = 0.48\textwidth]{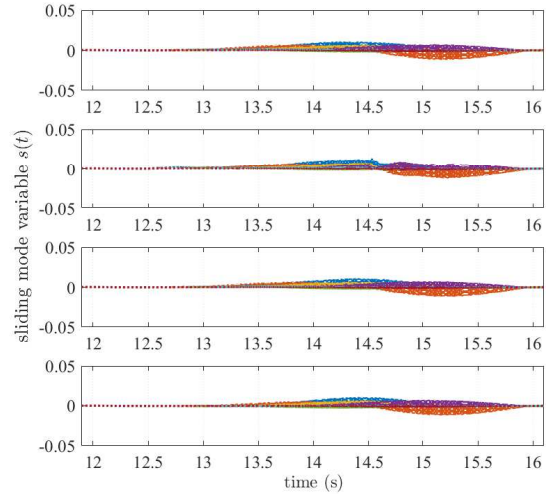}
\caption{The sliding mode variable $s(t)$ respectively for $\tau_{\mathrm{d}}^{\sin}(t)$, $\tau_{\mathrm{d}}^{\mathrm{sqt}}(t)$, $\tau_{\mathrm{d}}^{\mathrm{trg}}(t)$ and no disturbance (from top to bottom). Different colors represent different dimensions of $s(t)$.}
\label{fig:err}
\end{figure}

\begin{figure}[htbp]
\centering
\includegraphics[width = 0.48\textwidth]{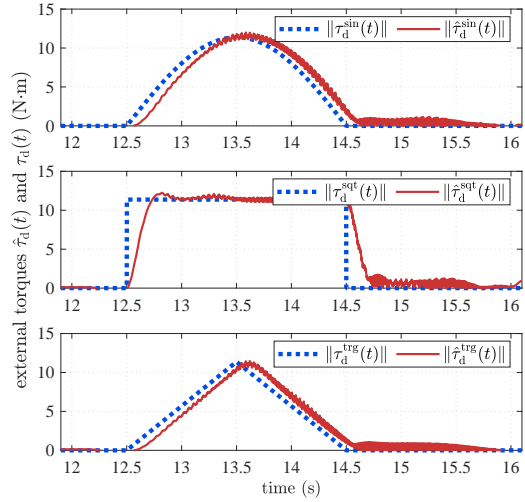}
\caption{The comparison between the predefined disturbances $\tau_{\mathrm{d}}(t)$ and the estimated results $\hat{\tau}_{\mathrm{d}}(t)$, respectively for $\tau_{\mathrm{d}}^{\sin}(t)$, $\tau_{\mathrm{d}}^{\mathrm{sqt}}(t)$ and $\tau_{\mathrm{d}}^{\mathrm{trg}}(t)$.}
\label{fig:dst}
\end{figure}

In this experiment, we exert three types of predefined disturbance torques on the commanded input interface of the robot during the motion of the manipulator, and make a comparison between the original disturbances and the estimated results. The three disturbance types are respectively the sinusoidal $\tau_{\mathrm{d}}^{\mathrm{sin}}(t)$, the square $\tau_{\mathrm{d}}^{\mathrm{sqr}}(t)$ and the triangle-form $\tau_{\mathrm{d}}^{\mathrm{trg}}(t)$,
\begin{equation*}
\begin{split}
\tau_{\mathrm{d}}^{\mathrm{sin}}(t) =&\, \sin \! \left( \frac{\pi}{2} (t - 12.5) \right) \! \tau_{\mathrm{t}},~12.5 \leq t \leq 14.5,\\
\tau_{\mathrm{d}}^{\mathrm{sqr}}(t) =&\, \tau_{\mathrm{t}},~12.5 \leq t \leq 14.5,\\
\tau_{\mathrm{d}}^{\mathrm{trg}}(t) =& \, \left\{ \begin{array}{ll}\!\!
(t-12.5) \tau_{\mathrm{t}}, & 12.5 < t \leq 13.5, \\
\!\!(t-12.5) \tau_{\mathrm{t}}, & 13.5 < t \leq 14.5,
\end{array} \right.
\end{split}
\end{equation*}
where $\tau_{\mathrm{t}} = [~6~~4.8~~3~~3.6~~4.2~~5.4~~1.2~]^{\top}\,$N$\cdot$m. For the default period $t \leq 12.5$ and $t > 14.5$, 
$\tau_{\mathrm{d}}^{\mathrm{sin}}(t) = \tau_{\mathrm{d}}^{\mathrm{sqr}}(t) = \tau_{\mathrm{d}}^{\mathrm{trg}}(t) = 0$. We apply these disturbances since they all resemble the waveform of collision forces in practice, and are frequently used in related literature~\cite{jiang2006fault, brambilla2008fault, edwards2000sliding, capisani2012manipulator}. Then, we implement the proposed observer (\ref{eq:observer}) to estimate each disturbance, and the parameters are determined as 
\begin{equation*}
\begin{split}
&L_1 \!=\! 156.7 I_7, ~L_2 \!=\! 2678 I_7, ~K_0 \!=\! 0.0585 I_7,~H \!=\! 0.2103 I_7,\\ 
&P_{11} \!=\! 24.55 I_7, ~P_{12} \!=\! -1.227I_7, ~P_{22} \!=\! 0.0718I_7, ~\delta_s \!=\! 0.05,
\end{split}
\end{equation*}
and $\varrho(e_{\zeta},\hat{\xi}) = 250+\|Q_{\zeta} e_{\zeta}\| + \|Q_{\xi} \hat{\xi}\|$, where $L$ and $P$ are solved using the MATLAB LMI toolbox, and $Q_{\zeta}$, $Q_{\xi}$ are calculated by $P$, $K$, and $A_L$ (See Theorem \ref{th:theo2}). The identified parameters $\hat{M}$, $\hat{N}$, $\hat{G}$ and $\hat{F}$ are online obtained from FRI, and $\bar{\dot{q}}$, $\dot{\hat{M}}$ are respectively obtained by differentiating the measured position $q(t)$ and the identified inertia matrix $\hat{M}$. The observer starts at an initial condition $\hat{x}(0) = 0$. The experimental results of the sliding mode variable $s(t)$ and the estimated disturbance $\hat{\tau}_{\mathrm{d}}(t)$ are respectively illustrated in Fig. \ref{fig:err} and Fig. \ref{fig:dst}. For brevity, we only present the results for $12 \leq t \leq 16$ since the rest period of time indicates similar results.

Fig. \ref{fig:err} shows that the sliding mode variable $s(t)$ is confined within the boundary-layer for all three types of disturbances, which confirms the estimation error convergence of the proposed observer (\ref{eq:observer}). Due to the system uncertainties during the motion of the manipulator, mainly the velocity error related terms $\tilde{N}(q, \dot{q}, \bar{\dot{q}})$ and $\tilde{F}(\dot{q}, \bar{\dot{q}})$, non-zero errors emerge between 13.5$\,$s and 15.5$\,$s, which is not the effect of the observer. This is also justified by comparing the results with the disturbance-free case (bottom of Fig. \ref{fig:err}) where similar non-zero errors also show up. Meanwhile, Fig. \ref{fig:dst} indicates that the proposed observer provides accurate estimation results for disturbances $\hat{\tau}_{\mathrm{d}}^{\sin}(t)$ and $\tau_{\mathrm{d}}^{\mathrm{trg}}(t)$. For $\hat{\tau}_{\mathrm{d}}^{\mathrm{sqt}}(t)$, however, obvious estimation errors are seen since the high-frequency partition of the original disturbance is filtered, which clearly shows the filter property of the proposed observer. The loss of the high-frequency partition of the time-dependent disturbance is the expense of the lack of its exact dynamics knowledge, which applies to nearly all the observer-based disturbance estimation methods. Therefore, the observer-based methods are only capable of reconstructing low-frequency disturbances, which, however, is sufficient to solve most practical problems.

\subsection{Experiment 2: Physical Contacts on the Robot}

\begin{figure*}[htbp]
	\centering
	\subfigure[Robot moving]{\label{fig:sgnsmp1}\includegraphics[height=0.22\textwidth]{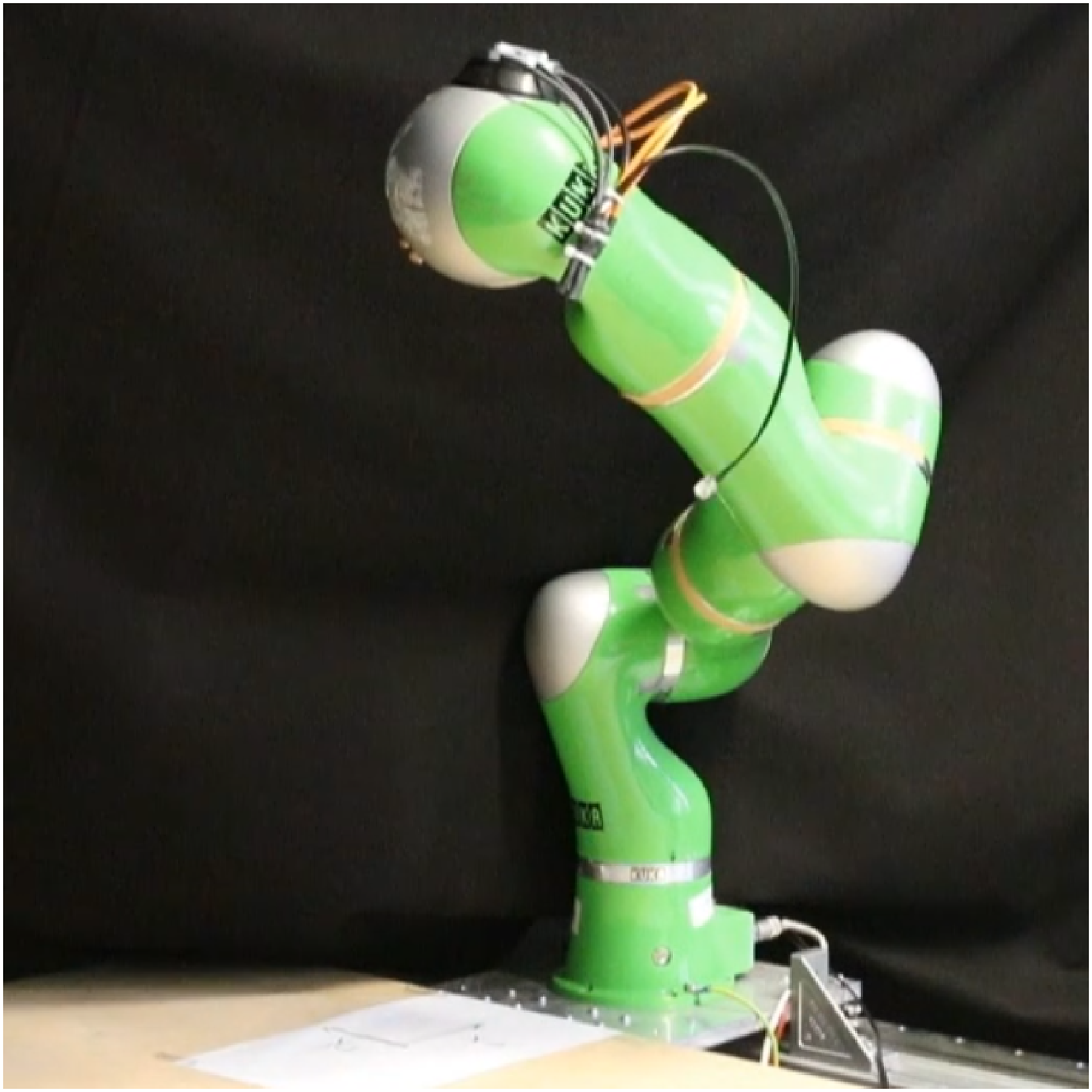}}		\hfill
	\subfigure[Contact 1]{\label{fig:sgnsmp2}\includegraphics[height=0.22\textwidth]{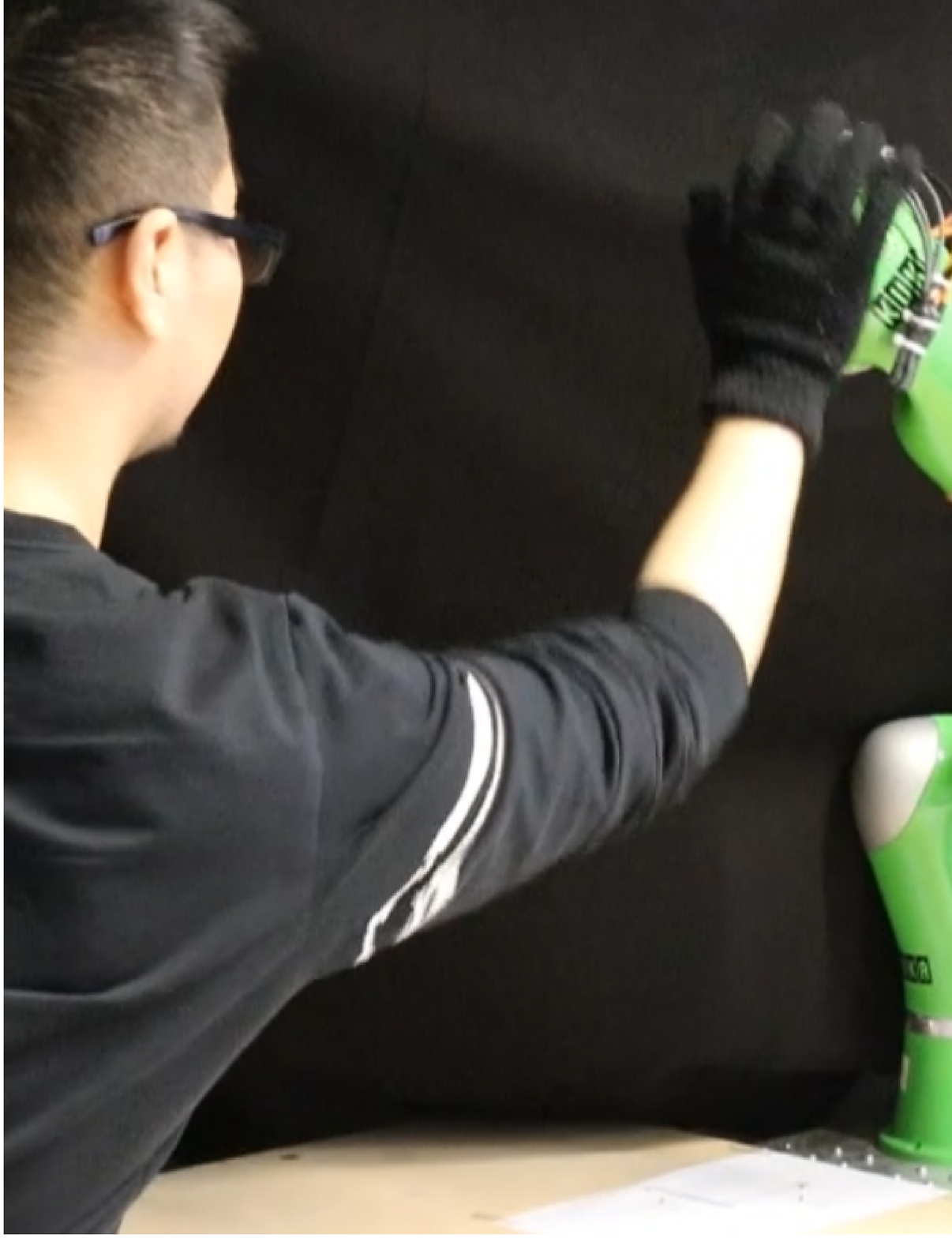}}	
    \hfill
	\subfigure[Contact 2]{\label{fig:sgnsmp3}\includegraphics[height=0.22\textwidth]{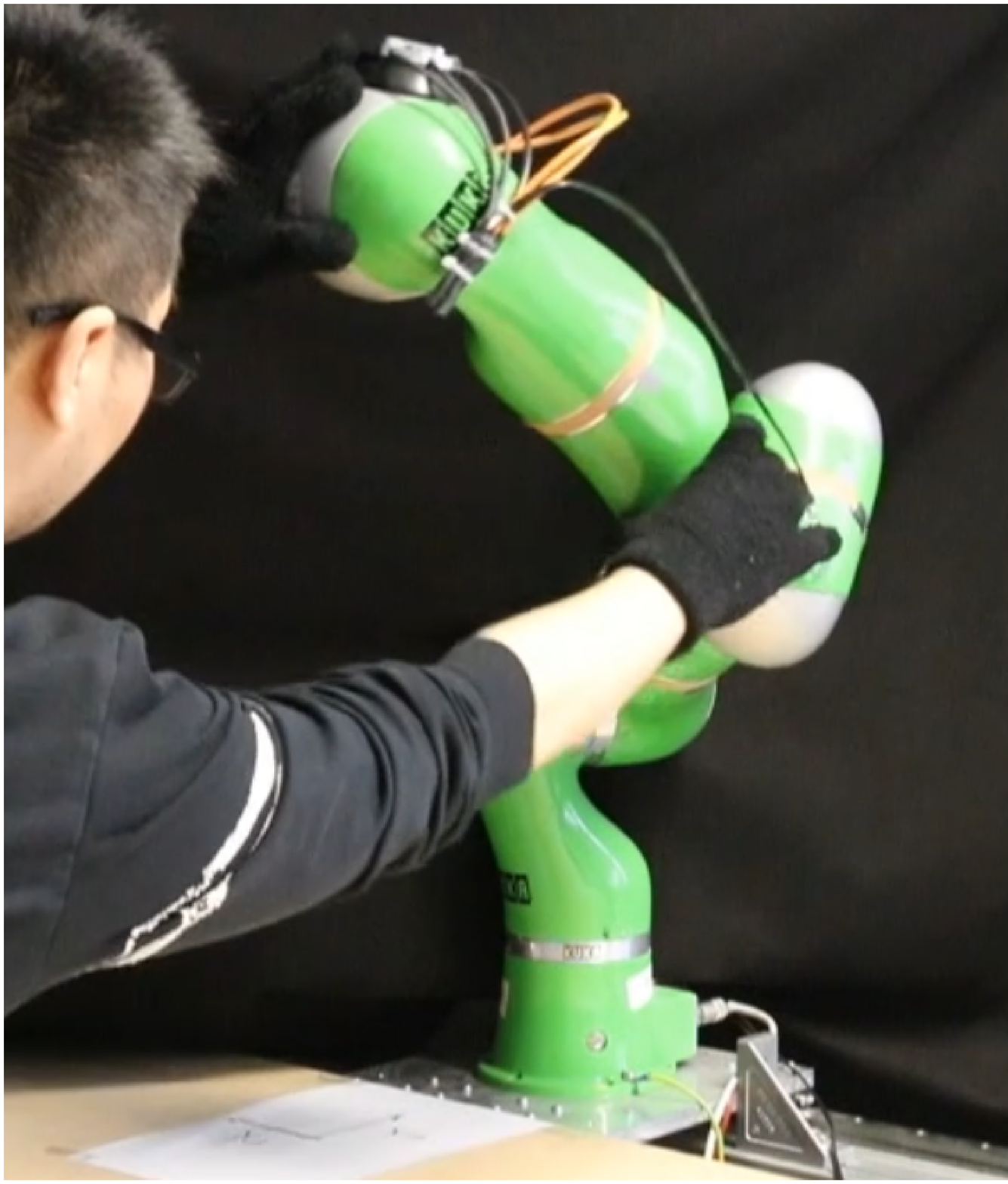}}	\hfill
        \subfigure[Contact 3]{\label{fig:sgnsmp4}\includegraphics[height=0.22\textwidth]{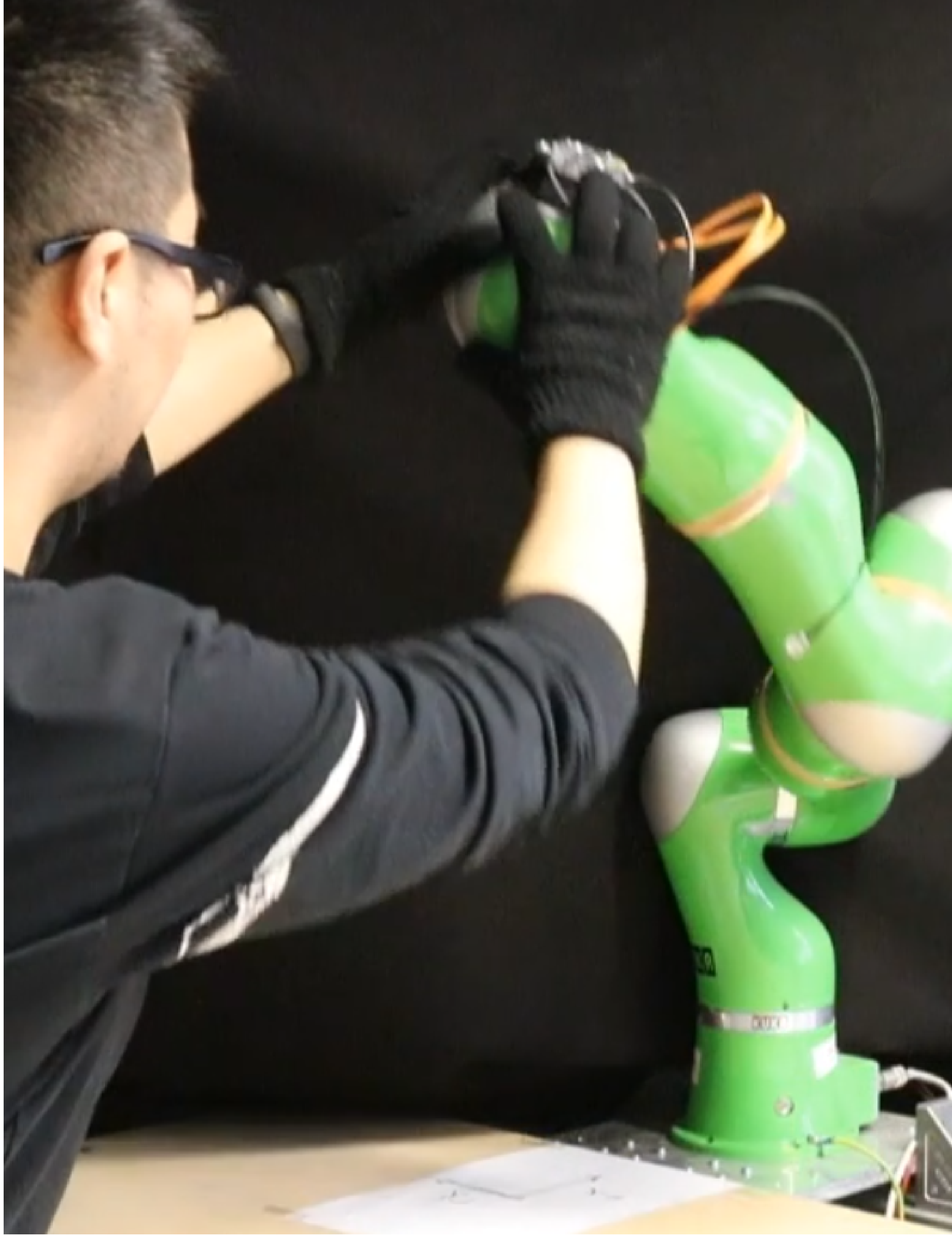}}	
	\caption{The method is tested during the movement of the robot. The robot is tested with various kinds of contact.}
\label{fig:sgnsmp}
\end{figure*}

\begin{figure}[htbp]
\centering
\includegraphics[width = 0.48\textwidth]{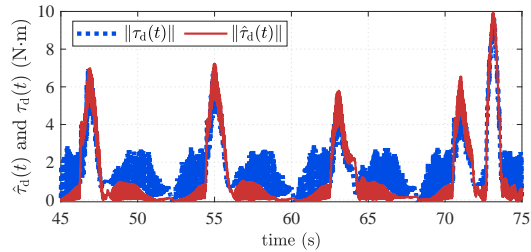}
\caption{The comparison between the estimated physical-contact torques $\hat{\tau}_{\mathrm{d}}(t)$ and the measured torques $\hat{\tau}_{\mathrm{d}}(t)$.}
\label{fig:trg}
\end{figure}

In this experiment, we make various physical contacts with the robot end-effector using a gloved hand, as shown in Fig.~\ref{fig:sgnsmp}, and compare the estimated torques $\hat{\tau}_{\mathrm{d}}(t)$ with the measurement $\tau_{\mathrm{d}}(t)$ obtained from the shaft torque sensors installed on the robot joints. The experimental results are shown in Fig. \ref{fig:trg}. To avoid redundancy, we only display the results for $45 \leq t \leq 75$.

Fig. \ref{fig:trg} reveals an accurate estimation result of the proposed method for the physical-contact torques, which is comparable to the measured values. It is also noticed that the estimated torques show a lower noise level than the sensory measurement, due to the filtering property of the observer. Therefore, from the results of the two experiments, we summarize that the proposed observer-based method provides accurate estimation results for external disturbances and applies to practical problems, such as contact-force estimation.

\section{Conclusion}\label{sec:concl}

In this paper, we present a novel observer-based disturbance estimation method for Euler-Lagrangian systems, which requires neither the inverse inertia matrix nor the system velocity feedback. Therefore, this method is quite promising to be applied to high-DoF robot manipulators for safe human-robot collaborative tasks. From the theoretical perspective, we provide a novel linearization scheme for Euler-Lagrangian systems and propose a new disturbance observer for second-order linear systems with partial state measurement. This is achieved by utilizing the partial dynamical collapse property and the equivalent control theory of sliding mode. Nevertheless, the feasibility of the proposed method on generic nonlinear systems still needs to be investigated, which is our main focus in future work.

\bibliographystyle{unsrt}  
\bibliography{references}

\end{document}